\documentclass[runningheads]{llncs}
\usepackage{graphicx}
\usepackage{amsmath}
\usepackage{amssymb}
\usepackage[section]{placeins}

\begin{document}
\title{Continual Learning of Feedback-based Molecular Communication}

\author{
Siddhant Setia\inst{1}
\and
Junichi Suzuki\inst{1}
\and
Tadashi Nakano\inst{2}
}
\authorrunning{S. Setia et al.}

\institute{
Department of Computer Science, University of Massachusetts, Boston\\
Boston, MA, 02125, USA\\
\email{Siddhant.Setia001@umb.edu}, 
\email{jxs@cs.umb.edu}\\
\and
Graduate School of Informatics, Osaka Metropolitan University\\
Osaka, Japan\\
\email{tnakano@omu.ac.jp}
}
\maketitle              

\begin{abstract}

This paper proposes and evaluates a new performance estimation method that leverages continual learning (CL) algorithms to carry out sequential simulation experiments for a feedback-based molecular communication protocol. As the protocol is sequentially examined in various experimental settings, the proposed CL-based performance estimators incrementally learn a series of unexperienced estimation tasks without compromising those that have been learned in the past. They are designed to work on a standard neural network architecture by customizing regularization and replay strategies in the loss function. Experimental results demonstrate that the proposed estimators can effectively learn on a continuous stream of simulation results and enhance the baseline neural network by improving estimation accuracy at a variety of computational costs. This paper's contribution is to establish the implications of CL in the field of molecular communication.

\end{abstract}

\section{Introduction}

Molecular communication research often advances in stepwise manners. When a communication protocol is studied, for example, it may be examined under the simplest setting first, such as in a one-dimensional noiseless space, to evaluate its characteristics and effectiveness. Then, it would be evaluated under more general or realistic settings gradually by introducing extra requirements or relaxing prior assumptions. The protocol may be examined in two- and then three-dimensional spaces by considering the presence of noise. While the transmitter~(Tx) and receiver~(Rx) bio-nanomachines may be placed close to each other first, the Tx-Rx distance would be gradually expanded, potentially with the aid of relay bio-nanomachines. Bio-nanomachines may be configured to be mobile eventually. 

As the complexity of these experimental settings increases, simulation studies are carried out in addition to analytical work. When they are computationally expensive, they are often augmented by machine learning (ML) methods. For example, if an ML model is built for a protocol, it is trained on prior simulation results to estimate a certain performance measure(s) such as latency and error rate. This can save computational cost to run extra simulations. 

As the protocol in question is studied in a stepwise manner, as described above, different ML models are built for different experimental settings. For example, an ML model is built for performance estimation in 1-D spaces. Then, a new ML model is built to examine the protocol in 2-D spaces by training the model with the  simulation results obtained in the 2-D spaces. While this 2-D ML model works for estimation tasks in 2-D spaces, it does not in 1-D spaces. To have a model that works in both 1-D and 2-D spaces, it needs to be trained on 1-D and 2-D simulation results. 
In other words, the model requires the ``re-training'' of 1-D results. To have a model for 1-D, 2-D and 3-D spaces, it has to be re-trained on 1-D and 2-D results. If re-training sessions are required repeatedly, they can cancel the benefit of saving computational cost with ML. 

In fact, it is possible to train an existing model for a new setting. For example, when transitioning from 1-D to 2-D spaces, a 1-D model can be trained on 2-D simulation results. However, the model ``forgets'' about 1-D spaces and yields poor estimation in the 1-D spaces, although it works in 2-D spaces. If the model is trained further with 3-D results, it forgets about 1-D spaces further and becomes even less capable for estimation tasks in 1-D spaces. This is called ``catastrophic forgetting''~\cite{mccloskey89catastrophic}. A challenge is to have an ML model acquire new knowledge from sequentially collected simulation results, integrate it with the current knowledge, and retain the previously-learned knowledge effectively.  

To address this challenge, this paper investigates a strategy to continuously adapt an ML model to new experimental settings in a stepwise manner and improve the model's ability to learn and solve sequential estimation tasks. The proposed strategy leverages continual learning (CL) methods, which are intended to mitigate the forgetting phenomena in ML~\cite{wang24cl}.

This paper considers a feedback-based protocol that enables in-sequence and timing-controlled molecular communication based on Stop-and-Wait Automatic Repeat Request (SW-ARQ)~\cite{burton72error},  assumes incremental learning scenarios where the protocol is examined in a sequence of different experimental settings, and customizes existing four CL methods for communication latency estimation in the scenarios. The CL methods are tailored to base a standard artificial neural network (ANN) and solve regression tasks. Experimental results demonstrate that the proposed CL-based performance estimators can effectively learn on a continuous stream of simulation results while alleviating forgetting phenomena. The proposed estimators enhance the baseline ANN by improving estimation accuracy at a variety of computational costs. They do not require the re-training of previously seen simulation results. 
This paper's contribution is to establish the implications of CL in the field of molecular communication. Findings of this paper can be applied to many other estimation/prediction tasks where forgetting phenomena can occur. 

\section{Background and Related Work}

Reliable in-sequence molecule delivery can play critical roles in various biomedical and healthcare applications. For artificial morphogenesis applications in regenerative medicine, bio-nanomachines made of living cells are designed to divide and grow to form three-dimensional multi-cellular structures such as tissues and organs. Molecular communication allows those bio-nanomachines to interact with each other using {\it artificial morphogens (AMs)}, which are modeled as the information molecules that encode morphological information. Reliable in-sequence molecule delivery is a critical foundation to control communication patterns (e.g., the order of propagated AMs and the interval of AM propagation) for adjusting the growth and differentiation of bio-nanomachines.

In drug delivery applications, therapeutic bio-nanomachines leverage natural molecular signals in the body, or molecular signals released by other bio-nanomachines, to identify target sites and release drug molecules, thereby avoiding/alleviating side-effects at non-target sites. Reliable in-sequence delivery allows bio-nanomachines to modulate tumor behavior by controling the order and duration of drug molecule presence. 

In order to enhance the reliability of molecular communication, several feedback-based protocols have been studied. Feedback-based rate control schemes are proposed for diffusive transports in~\cite{nakano13jsac,felicetti14tcpmol}. In-sequence and at-least-once molecule delivery schemes are examined with SW-ARQ protocols in~\cite{wang14arq,bai15arq-bacteria,singh23ecc,mitzman15swarq}. Diffusive transports are considered in~\cite{wang14arq,bai15arq-bacteria,singh23ecc}, while diffusive, directional and diffusive-directional hybrid transports are considered in~\cite{mitzman15swarq}. None of these feedback-based protocols are backed by ML; this paper is the first attempt for performance estimation with SW-ARQ in molecular communication. 

Various ML methods have been studied for molecular communication; e.g.~\cite{cheng2025bio,cheng25dl,lee2017mlchannel,yilmaz2017ml,casaleiro2024sync,cheng2025localtrack,kara2022idx}. Most of them perform regression tasks, as in this paper, while a classification task is considered in~\cite{kara2022idx}, which uses a convolutional NN (CNN) to predict the source of each information molecule for molecular index modulation. This paper is similar to~\cite{cheng2025bio,cheng25dl}, which estimate the arrival time of propagated molecules at the Rx in a fixed (diffusive) transport. 
A transformer NN is used for a 2-D noiseless environment in~\cite{cheng2025bio}, and a multi-layer perceptron is used for a 3-D noiseless environment in~\cite{cheng25dl}. In contrast, this paper considers variable noise levels and transport choices in a 3-D environment and uses CL-enhanced NNs to estimate the latency of molecule propagation from the Tx and its feedback from the Rx. 
In~\cite{lee2017mlchannel,yilmaz2017ml}, NNs are used to estimate the fraction of molecules that arrive at the Rx by a certain deadline in 3-D noiseless environments. In~\cite{casaleiro2024sync}, a CNN and a recurrent NN (RNN) are used to predict the type of information encoded in a molecule with a 3-D noisy environment. In~\cite{cheng2025localtrack}, the location of a mobile Tx is estimated with a transformer in a 2-D noiseless environment. None of these ML-based work employs CL; their ML models require re-training or suffer forgetting phenomena if they are incrementally trained on different experimental settings. To the best of the authors' knowledge, this paper is the first attempt to investigate CL in molecular communication.

\section{Molecular Communication Model}
\label{sec:molcom}

\begin{figure*}[th]
  \centering
	\resizebox{1.0\linewidth}{!}{\includegraphics{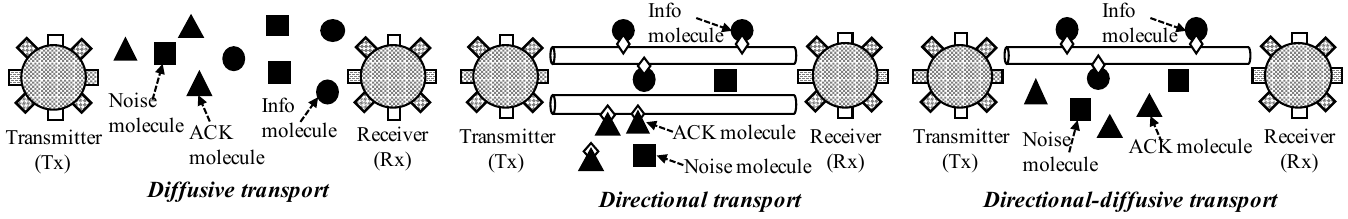}}

	\caption{Diffusive, Directional and Hybrid Transports with SW-ARQ}
	\label{fig:transports}
\end{figure*}

\begin{figure}[th]
  \centering
	\resizebox{0.7\linewidth}{!}{\includegraphics{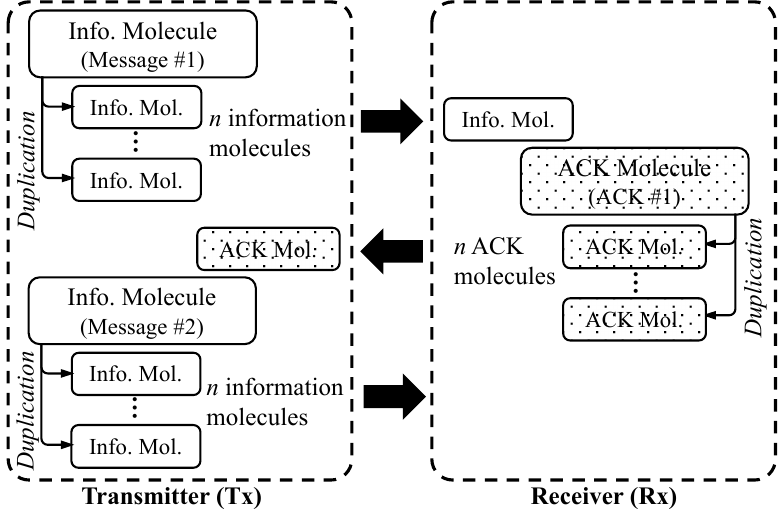}}
	\caption{Interaction between the Tx and the Rx}
	\label{fig:txrx}
\end{figure}

This paper assumes a bounded three-dimensional aqueous environment where the stationary transmitter (Tx) and receiver (Rx) bio-nanomachines exchange information and acknowledgement (ACK) molecules. Fig.~\ref{fig:transports} overviews diffusive, directional and hybrid transports for molecules to travel between Tx and Rx. The Tx releases information molecules, each of which encodes a certain message and carries it to the Rx (Fig.~\ref{fig:txrx}). The Rx is assumed to capture an information molecule when the molecule has a physical contact with the Rx. 
The Rx releases ACK molecules, each of which acknowledges the message. The Tx is assumed to capture an ACK molecule when it contacts the Tx's surface. Noise molecules are stationary and interfere (i.e., collide) with other molecules.

In a diffusive transport, a molecule diffuses via random thermal motion, which is governed by the diffusion coefficient~$D$ in each dimension: $D = \partial x^2 / (2 \times \partial t)$ where $x$ denotes the distance of molecular movement during time~$t$ on a particular dimension. When a molecule collides with another molecule, it randomly moves to another position with $D$.  

A directional transport utilizes a microtubule between Tx and Rx. Based on some findings in biomedical engineering (e.g.,~\cite{vale96nature}), the microtubule is configured to directionally guide molecular motors (e.g., kinesin, dynein and myosin). Each information/ACK molecule is attached to a molecular motor, and it travels at a constant velocity for an expected distance and diffuses away. It also diffuses away when it collides with another molecule. While diffusing, a molecule with a molecular motor may contact the microtubule and begin to travel on it again. 

A hybrid transport uses both diffusive and directional transports. This paper deploys a directional transport for information molecules and a diffusive transport for ACK molecules.

This paper employs SW-ARQ for molecular transmissions. The Tx releases $n$ duplicated information molecules that encode a particular message (e.g., Message 1 in Fig.~\ref{fig:txrx}). Once the Rx receives one of them, it releases $n$ duplicated ACK molecules. It does not respond to subsequent arrivals of the information molecules that encode the same message. 
Upon receiving one of the ACK molecules, the Tx releases $n$ information molecules that encode the next message (e.g., Message 2 in Fig.~\ref{fig:txrx}). It does not respond to subsequent arrivals of the ACK molecules for the previous message. 

If the Tx does not receive an ACK molecule 
for a message in a certain interval, called retransmission timeout interval (RTO), it retransmits another set of $n$ information molecules for the message. Molecule retransmission may occur multiple times until the Tx receives an ACK molecule or exceeds a predefined number of retransmissions. 
If the Rx does not receive an information molecule that encodes the $i$-th message within RTO after sending ACK molecules for the ($i-$1)-th message, it retransmits $n$ ACK molecules for the previous message.

\section{Incremental Performance Estimation}

This paper examines the 
protocol described in Section~\ref{sec:molcom} through 
performance estimation tasks. Each estimation task defines a set of experimental settings including environmental parameters (e.g., the environment size, noise level and Tx-Rx distance) and communication parameters (e.g., transport choice, the number of duplicated information/ACK molecules and RTO). Each task focuses on the roundtrip time (RTT) for a message delivery and estimates it under given experimental settings. RTT is the time required for the Tx to receive an ACK molecule for a message since releasing information molecules for the message.

\begin{figure*}[th]
  \centering
	\resizebox{1.0\linewidth}{!}{\includegraphics{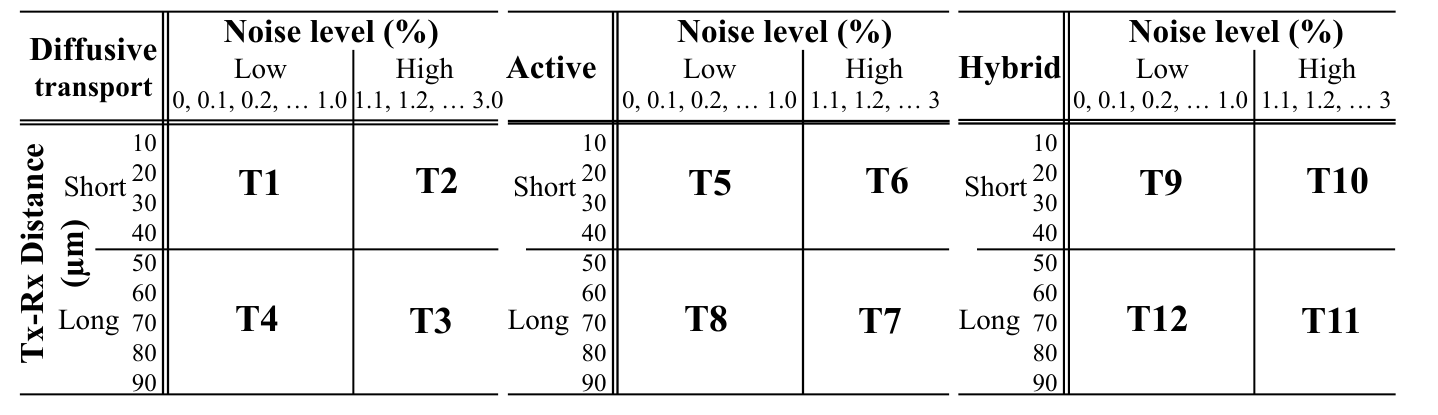}}
	\caption{Sequential Tasks (T1 to T12) for RTT Estimation}
	\label{fig:tasks}
\end{figure*}

The RTT estimation model is designed to take a set of experimental settings as an input and outputs an estimated RTT value. The model is trained incrementally through 12 tasks shown in~Fig.~\ref{fig:tasks}. Each task runs a series of simulations under the current experimental settings and feeds simulation results to the model to perform a supervised training session. For example, in the first task (T1), simulations are carried out in the diffusive transport for each pair of Tx-Rx distance and noise level to obtain RTT values. The Tx-Rx distance, noise level and other experimental settings are used as a training input, and the median RTT is used as the true output value. A pair of a training input and a true output forms a training data sample. A number of training data samples are generated through simulations and fed to the model to learn about T1.

In each training session, the model has no access to the training data from prior tasks. At test time, the model is used to perform RTT estimation for the current and all prior tasks. It does not know which task
a given test input belongs to. For example, when the model is trained for the second task (T2), it cannot use the training data from T1 but can use the one generated in T2 only. In testing, the model is used to estimate RTT for both T1 and T2.

Following the terminology framework in~\cite{vandeven22incremental}, this paper studies {\it domain-incremental} CL where an SW-ARQ protocol is sequentially evaluated in 
various experimental settings (or domains) and its RTT estimation model is incrementally trained on the  training data generated in a sequence of distinct tasks. 

\section{Continual Learning Algorithms}

This paper examines four CL algorithms: Learning Without Forgetting (LWF)~\cite{li18lwf}, Elastic Weight Consolidation (EWC)~\cite{kirkpatrick17ewc}, Continual Learning for Regression Tasks (CLeaR)~\cite{he21clear}, and Dark Experience Replay (DER)~\cite{buzzega20der}. They are implemented on top of a common baseline NN. They extend the baseline loss function to perform regularization and replay strategies for CL. Since LWF, EWC and DER are designed for classification tasks, this paper customizes their loss functions to perform regression tasks (RTT estimation tasks). 

\subsection*{Baseline}

The baseline NN is a standard 2-layer feedforward network with Rectified Linear Unit (ReLU) as an activation function. Its first layer takes 12 input values and has 20 hidden nodes. Its second layer linearly reduces 20 values to a single output value (i.e., RTT value). Input and output values are normalized into [0, 1] with min-max normalization. The baseline NN uses the following loss function: 

\begin{equation*}
\mathcal{L}_{\text{baseline}}(\Theta) = \mathcal{L}_{\text{mse}}(y(\Theta)^K - \hat{y}^K)
\end{equation*}

$\Theta$ denotes a set of model parameters, weights and biases, which are to be adjusted with a stochastic gradient descent method in a training session. ${y}(\Theta)^K$ is the model's output (predicted RTT value) in the current $K$-th task. $\hat{y}^K$ is the true output value (true RTT value). $ \mathcal{L}_{\text{mse}}$ computes the mean squared error (MSE) between ${y}(\Theta)^K$ and  $\hat{y}^K$. 

\subsection*{Learning Without Forgetting (LWF)}

LWF's loss function is defined as a linear weighted sum of a loss term and a regularization term~\cite{li18lwf}: 

\begin{equation*}
\mathcal{L}_{\text{LWF}}(\Theta) = (1 - \lambda) \mathcal{L}_{\text{cross}}\left( y(\Theta)^K - \hat{y}^K \right) + \frac{\lambda}{K-1} \sum_{k=1}^{K-1} \mathcal{L}_{\text{kdl}} \left( y(\Theta)^k - \hat{y}(\Theta_{\text{old}})^k \right)
\label{eq:lwfinitial}
\end{equation*}

The first term computes the loss in the current $K$-th task with a standard cross-entropy loss function: $\mathcal{L}_{\text{cross}}$. The second term is for functional regularization, which discourages to significantly change the input-output mapping that have been learned in the previous $K-1$ tasks. $\lambda$ is the regularization strength parameter to control the weights for the loss and regularization terms. 

$\mathcal{L}_{\text{kdl}}$ is the knowledge distillation loss function~\cite{hinton2015distillingknowledgeneuralnetwork}. 
$y(\Theta)^k$ denotes the output that the model generates for a previous $k$-th task ($1 \leq k \leq K-1$) with the current model parameters: $\Theta$. $y(\Theta_{\text{old}})^k$ denotes the output that the model generated for a previous $k$-th task with the model parameters that were obtained in the $k$-th task: $\Theta_{\text{old}}$. 

Since LWF is designed for classification tasks, this paper customizes its loss function by replacing $\mathcal{L}_{\text{cross}}$ and $\mathcal{L}_{\text{kdl}}$ with $\mathcal{L}_{\text{mse}}$: 

\begin{equation*}
\mathcal{L}_{\text{LWF}}(\Theta) = (1 - \lambda) \mathcal{L}_{\text{mse}}\left( y(\Theta)^K - \hat{y}^K \right) + \frac{\lambda}{K-1} \sum_{k=1}^{K-1} \mathcal{L}_{\text{mse}} \left( y(\Theta)^k - \hat{y}(\Theta_{\text{old}})^k \right)
\label{eq:lwffinal}
\end{equation*}

\subsection*{Elastic Weight Consolidation (EWC)}

EWC's loss function is also defined as a linear weighted sum of a loss term and a regularization term~\cite{kirkpatrick17ewc}: 

\begin{equation*}
\label{eq:ewcinitial}
\mathcal{L}_{\text{EWC}}(\Theta) = \mathcal{L}_{\text{mse}}\left( y(\Theta)^K - \hat{y}^K \right) +  \frac{\lambda}{2} \sum_{k=1}^{K-1} \left( \sum_{i=1}^{N_{\text{param}}} F_{i}^{k} \left( \Theta_{i} - \hat{\Theta}_{i}^{k} \right)^2 \right)
\end{equation*}

The second term is for parameter regularization, which penalizes the changes of the model parameters $\Theta$ that are important for the previous $K-1$ tasks. $N_{\text{params}}$ denotes the total number of model parameters. $\Theta_i$ is the value of the $i$-th parameter at the end of the training session in the current $K$-th task. $\hat{\Theta}_i^{k}$ is the value of the $i$-th parameter at the end of the training session in a previous $k$-th task. $F_i^k$ denotes the estimated importance of the $i$-th parameter in the $k$-th task. It is calculated as the $i$-th diagonal element of the Fisher information matrix in the $k$-th task~\cite{kirkpatrick17ewc}. In this paper, it is approximated as the variance of 
$\dfrac{\partial \mathcal{L}_{\text{mse}}(y(\Theta)^k - \hat{y}^k)}{\partial \Theta_i}$
over all training data samples in the $k$-th task.

\subsection*{Continual Learning for Regression Tasks (CLeaR)}

CLeaR customizes EWC's loss function as follows~\cite{he21clear}:

\begin{equation*}
\label{eq:clearloss}
\mathcal{L}_{\text{CLeaR}}(\mathbf{\Theta}) = \mathcal{L_{\text{mse}}}\left( y(\Theta)^K - \hat{y}^K \right) + \frac{\lambda}{2} \sum_{i=1}^{N_{\text{param}}}  F_{i}^{K-1} \left( \Theta_{i} - \hat{\Theta}_{i}^{K-1} \right)^2
\end{equation*}

CLeaR computes  $F_i^k$ only for the most recent task; i.e., ($K$-1)-th task. A hyperparameter $\alpha$ and a minimum value of MSE are defined before training and a threshold is calculated using, 

\begin{equation}
\label{eq:clearThreshold}
\text{threshold} = \alpha MSE
\end{equation}

For a task $K$ and a predicted value $y(\Theta)^K$, an MSE loss is computed. If the loss exceeds a threshold, the input and true value are added to a novelty buffer. Otherwise, they are added to a familiarity buffer. When the total buffer size reaches a limit, which is defined while training, the model stops training on the current task and retrains on the novelty buffer. It is then tested on the familiarity buffer and if any sample has a lower MSE loss than the current minimum MSE, the minimum MSE and threshold are updated with this value. The novelty and
familiarity buffers are updated based on model predictions $y(\Theta)^K$.

\subsection*{Dark Experience Replay (DER)}

DER's loss function is defined as a linear weighted sum of a loss term and two regularization terms~\cite{buzzega20der}: 

\begin{equation*}
\begin{split}
\label{eq:derpp}
\mathcal{L}_{\text{DER}}(\Theta) = \mathcal{L}\left( y(\Theta)^K - \hat{y}^K \right) &+ \alpha \mathbb{E}_{(x', y', z') \sim \mathcal{M}} [\|z' - h_{\Theta}(x')\|_2^2] \\
&+ \beta \mathbb{E}_{(x'', y'', z'') \sim \mathcal{M}} [\mathcal{L}(y'' - f_{\Theta}(x''))]
\end{split}
\end{equation*}

The second and third terms are for functional regularization, which discourages to significantly change the input-output mapping that have been learned in the past. $\alpha$ and $\beta$ control the weights of the regularization terms. 

To calculate the regularization terms, DER samples the past training data into a buffer~$\mathcal{M}$ and replays a randomly-selected subset of them. $(x', y', z')$ and $(x'', y'', z'')$ are training data samples drawn from~$\mathcal{M}$. $x'$ and $x''$ denote input values. $y'$ and $y''$ denote the true output values. $z'$ and $z''$ denote ``logits." Logits is the raw, unnormalized output that the model generates right before applying softmax in the last layer. $f_\theta(x)$ and  $h_\theta(x)$ denote the final output and logits that the model generates with the current parameters~$\Theta$. $f_\theta(x) \triangleq \text{softmax}(h_\theta(x))$. 

Since DER is designed for classification tasks, this paper customizes its loss function by replacing loss terms with $\mathcal{L}_{\text{mse}}$: 

\begin{equation*}
\begin{split}
\label{eq:derpp}
\mathcal{L}_{\text{DER}}(\Theta) = \mathcal{L}_{\text{mse}}\left( y(\Theta)^K - \hat{y}^K \right) &+ \alpha \mathbb{E}_{(x', y', z') \sim \mathcal{M}} [\mathcal{L}_{\text{mse}} (z' - h_{\Theta}(x'))] \\
&+ \beta \mathbb{E}_{(x'', y'', z'') \sim \mathcal{M}} [\mathcal{L}_{\text{mse}} (y'' - f_{\Theta}(x''))]
\end{split}
\end{equation*}

\section{Experimental Results}

This section evaluates the proposed CL estimators through Monte-Carlo simulations (particle-based simulations). Table~\ref{tab:param-mc} shows the parameter settings for molecular communication. Each training and test data sample is prepared with the median RTT obtained from 500 independent simulations. 

\begin{table}[htb]
\centering
\caption {Parameter Settings for Molecular Communication}
\vspace{-0.25\baselineskip}
\begin{tabular}{|l||c|}
\hline
{\bf Parameter} & {\bf Value} \\
\hline \hline 
Size length of the environment & 150 $\mu$m \\
\hline 
Diameter of Tx and Rx & 5 $\mu$m \\
\hline
Tx to Rx distance & 10 to 90 $\mu$m \\
\hline 
Diameter of an information/ACK molecule & 1 $\mu$m \\
\hline 
Diameter of a noise molecule & 1 $\mu$m \\
\hline
Number of noise molecules & 0 to 10$^5$   \\
\hline 
Diffusion coefficient ($D$) & 0.5 \\
\hline 
Velocity of a molecular motor on a microtubule   & 1 $\mu$m/s \\
\hline
\begin{tabular}{@{}l@{}}Expected travel distance of\\a molecular motor on a microtubule\end{tabular} & 4 $\mu$m \\
\hline
\# of duplicated molecules per message/acknowledgement & 10 \\
\hline
Max number of molecule retransmissions & 5 \\
\hline 
\end{tabular}
\label{tab:param-mc}
\end{table}

The boundaries of the environment are simulated in a non-replusive manner. Molecules do not rebound against boundaries. Molecule-to-molecule collisions are also simulated in a non-replusive manner. Each molecule is prohibited to move to the location of another molecule. Colliding molecules do not rebound with each other. All simulations are carried out to perform a single message transmission, in which duplicated information molecules travel from Tx to Rx and duplicated acknowledgement molecules travel from Rx to Tx.

All CL and baseline algorithms use the following hyperparameter values: 

\begin{itemize}
  \item {Epochs}: 100. This determines the number of iterations to adjust the model parameters (weights and biases) by feeding training data to the model.
  \item {Batch size}: 128. This determines the number of training data samples that pass forward and backward through the neural network in each epoch. 
  \item {Learning rate}: 0.001. This controls how much to adjust the model parameters in each epoch. 
\end{itemize}

Table~\ref{tab:train-hyper-params} shows the hyperparameter values specific to CL algorithms. They are chosen through grid search over parameter combinations. 

\begin{table}[htb]
    \centering
    \caption{CL Hyperparameters}
    \begin{tabular}{|l||l|c|}
    \hline
    \textbf{Algorithm} & \textbf{Parameter} & \textbf{Value} \\
    \hline
    \hline
    LWF & $\lambda$ & 0.9 \\
        & Early stopping patience & 10 \\
    \hline
    EWC & $\lambda$ & 0.75 \\
        & Early stopping patience & 10 \\
    \hline
    CLeaR & $\lambda$ & 2 \\
          & $\alpha$ & 0.5 \\
          & Buffer Size & 50 \\
          & Early stopping patience & 10 \\
    \hline
    DER & $\alpha$ & 200 \\
        & $\beta$ & 200 \\
        & Buffer size & 5 \\
        & Early stopping patience & 10 \\
    \hline
    \end{tabular}
    \label{tab:train-hyper-params}
\end{table}

By using the aforementioned parameter settings, individual estimators are configured to perform RTT estimation for the 12 tasks in Fig.~\ref{fig:tasks}. In each test run, those tasks are sequentially arranged to form 12 different evaluation scenarios. The first scenario runs the first task (T1 in Fig.~\ref{fig:tasks}) only. The second scenario sequentially runs the first and second tasks (T1 and T2 in Fig.~\ref{fig:tasks}). The last  scenario runs all 12 tasks from T1 to T12 sequentially. In each task, each CL estimator is tested with a test dataset specific to the task. Its test error is measured as the MSE between estimated and true RTT values: $\mathcal{L_\text{mse}}$. Through the entire set of evaluation scenarios, the following matrix of test errors is produced for each CL estimator: 

\[
\begin{bmatrix}
  \mathcal{L_\text{mse}}_1^1 \\
  \mathcal{L_\text{mse}}_1^2 & \mathcal{L_\text{mse}}_2^2 \\
  \mathcal{L_\text{mse}}_1^3 & \mathcal{L_\text{mse}}_2^3 & \mathcal{L_\text{mse}}_3^3 \\
  \vdots & \vdots & \vdots & \ddots \\
  \mathcal{L_\text{mse}}_1^{K-1} & \mathcal{L_\text{mse}}_2^{K-1} & \mathcal{L_\text{mse}}_3^{K-1} & \cdots & \mathcal{L_\text{mse}}_{K-1}^{K-1} \\
  \mathcal{L_\text{mse}}_1^K & \mathcal{L_\text{mse}}_2^K & \mathcal{L_\text{mse}}_3^K & \cdots & \mathcal{L_\text{mse}}_{K-1}^K & \mathcal{L_\text{mse}}_K^K
\end{bmatrix}
\]

The rows of the matrix represent evaluation scenarios, and its columns represent RTT estimation tasks. For example, $\mathcal{L_\text{mse}}_3^K$ denotes the test error at the third task (T3) in the $K$-th scenario. $K$ denotes the total number of tasks. $K$ = 12 in this paper.  

To evaluate and compare individual estimators in terms of RTT estimation accuracy, this paper uses the following metrics. 

\begin{equation*}
\label{eq:duringtrain}
\mathcal{L}_{\text{plasticity}} = \frac{1}{K}\sum_{k=1}^{K} \mathcal{L_\text{mse}}_k^k
\end{equation*}

\begin{equation*}
\label{eq:aftertrain}
\mathcal{L}_{\text{stability}} = \frac{1}{K}\sum_{k=1}^{K} \mathcal{L_\text{mse}}_k^K
\end{equation*}

$\mathcal{L}_{\text{plasticity}}$ is the average of diagonal test errors in the above matrix. Each of them ($\mathcal{L_\text{mse}}_k^k$) indicates the test error in the last task in a particular scenario (i.e., the most recent $k$-th task in the $k$-th scenario). $\mathcal{L}_{\text{stability}}$ considers the test errors of all tasks in the last ($K$-th) scenario in comparison. Lower $\mathcal{L}_{\text{plasticity}}$ and $\mathcal{L}_{\text{stability}}$ values mean higher estimation accuracy. The gap between $\mathcal{L}_{\text{plasticity}}$ and $\mathcal{L}_{\text{stability}}$ increases as forgetting phenomena occur through incremental learning of estimation tasks.
$\mathcal{L}_{\text{plasticity}} = \mathcal{L}_{\text{stability}}$ if no forgetting phenomena occur.

Table \ref{tab:avgmse-stddev} shows the average $\mathcal{L}_{\text{plasticity}}$ and $\mathcal{L}_{\text{stability}}$ of individual estimators over 20 independent test runs. In both metrics, CL estimators consistently yield lower estimation errors than the baseline estimator. LWF is found the most accurate as its error is the lowest in both metrics. CLeaR yields the second best accuracy in both metrics. The baseline is the least accurate estimator. The error differences between LWF and the baseline are 0.0183 and 0.02276 in  $\mathcal{L}_{\text{plasticity}}$ and $\mathcal{L}_{\text{stability}}$, respectively. Approximately, the MSE difference of 0.01 results in the RTT difference of 1,500 seconds (25 minutes).

$\mathcal{L}_{\text{stability}}$ is higher than $\mathcal{L}_{\text{plasticity}}$ with all estimators. This is reasonable because they should forget about previously learned tasks to some extent as they are trained on new tasks. The baseline suffers the highest increase rate from $\mathcal{L}_{\text{plasticity}}$ to $\mathcal{L}_{\text{stability}}$: 6.5$\%$. Forgetting phenomena occur in the baseline the most significantly among all estimators. CLeaR yields the lowest increase rate: 2.9$\%$. 

\begin{table}[htb]
    \centering
    \caption{Average $\mathcal{L}_{\text{plasticity}}$ and $\mathcal{L}_{\text{stability}}$ and their standard deviations over all tasks}
    \begin{tabular}{|l||c|c|c|}
    \hline
    \textbf{Algorithm} & \textbf{$\mathcal{L}_{\text{plasticity}}$} & \textbf{$\mathcal{L}_{\text{stability}}$} & \textbf{Increase Rate (\%)} \\
    \hline
    \hline
    Baseline & 0.12517 $\pm$ 0.00023 & 0.13307 $\pm$ 0.00188 & 6.5 \\
    LWF & 0.10687 $\pm$ 0.00073 & 0.11031 $\pm$ 0.00107 & 3.2\\
    EWC & 0.12100 $\pm$ 0.00068 & 0.12757 $\pm$ 0.00110 & 5.4\\
    CLeaR & 0.10977 $\pm$ 0.00057 & 0.11299 $\pm$ 0.00082 & 2.9 \\
    DER & 0.11182 $\pm$ 0.00305 & 0.11846 $\pm$ 0.00102 & 5.9\\
    \hline
    \end{tabular}
    \label{tab:avgmse-stddev}
\end{table}

Table~\ref{tab:cl-times} shows the computational costs for individual estimators to perform their training sessions. Total time means the time required to incrementally train an estimator on all tasks. Per-task time is the average training time per task. 
As in~Table~\ref{tab:cl-times}, two CL estimators (CLeaR and DER)  spend shorter training time than the baseline. LWF's training time is the longest due to its computational complexity, while it yields the highest estimation accuracy (Table~\ref{tab:avgmse-stddev}). 

 \begin{table}[htb]
    \centering
    \caption{Computational cost for training}    
    \begin{tabular}{|l||r|r|r|}
    \hline
    \textbf{Algorithm} & \multicolumn{2}{c|}{\textbf{Training (sec)}} & \textbf{Standard Deviation} \\
    \cline{2-3}
    & Total Time & Per-task Time & \\
    \hline
    \hline
    Baseline & 13.4898 & 1.1241 & 0.2216 \\
    LWF & 387.9389 & 32.3282 & 37.6129 \\
    EWC & 16.7663 & 1.3971 & 2.3278 \\
    CLeaR & 6.9737 & 0.5811 & 1.0999 \\
    DER & 10.5451 & 0.8787 & 2.0418 \\
    \hline
    \end{tabular}
    \label{tab:cl-times}
\end{table}

\begin{figure}[htb]
    \centering
    \includegraphics[width=0.75\textwidth]{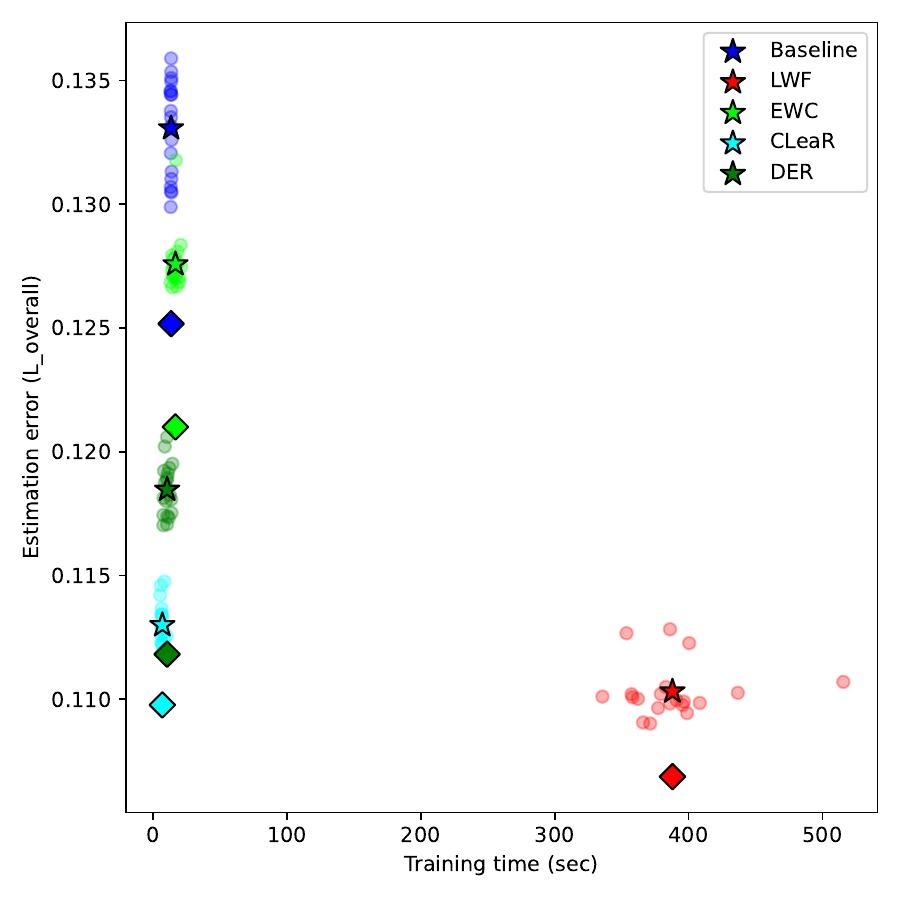}
    \caption{Trade-off between estimation accuracy and computational cost}
    \label{fig:error_time}
\end{figure}

Fig.~\ref{fig:error_time} illustrates the trade-off between estimation accuracy and computational cost. A circular dot represents $\mathcal{L}_{\text{stability}}$ and total training time on a particular test run. A star and a diamond indicate the average $\mathcal{L}_{\text{stability}}$ and average total training time over 20 test runs, respectively.  As mentioned earlier, LWF's average $\mathcal{L}_{\text{stability}}$ is the lowest among all estimators, while its average training time is the longest among all estimators. Its standard deviation of training time is the highest as well among all estimators (Table~\ref{tab:cl-times}). 

The baseline's average $\mathcal{L}_{\text{stability}}$ is the worst among all estimators, and its standard deviation of $\mathcal{L}_{\text{stability}}$ is also the worst (Tables~\ref{tab:avgmse-stddev}). In comparison to the baseline, CLeaR and DER yield lower $\mathcal{L}_{\text{stability}}$ and shorter training time on average. CLeaR is the second best in the average $\mathcal{L}_{\text{stability}}$ and the best in the average training time (Tables~\ref{tab:avgmse-stddev} and~\ref{tab:cl-times}). DER is the third best in the average $\mathcal{L}_{\text{stability}}$ and the second best in the average training time. 

To analyze how forgetting phenomena occur and vary in a sequence of estimation tasks, this paper employs the following metric: 

\begin{equation*}
\label{eq:forgetting-ratio}
\text{Forgetting Ratio ($F_{r}$)} = \frac{1}{K} \sum_{k=1}^{K} \frac{max(0, \mathcal{L_\text{mse}}_k^K - \mathcal{L_\text{mse}}_k^k)}{\mathcal{L_\text{mse}}_k^k}
\end{equation*}

$F_{r}$ is essentially the average ratio of difference between $\mathcal{L}_{\text{stability}}$ and $\mathcal{L}_{\text{plasticity}}$ over $\mathcal{L}_{\text{plasticity}}$ in $K$ tasks. It is computed for each estimator by varying $K$ from 2 to 12.

\begin{figure}[htb]
    \centering
    \includegraphics[width=1\textwidth]{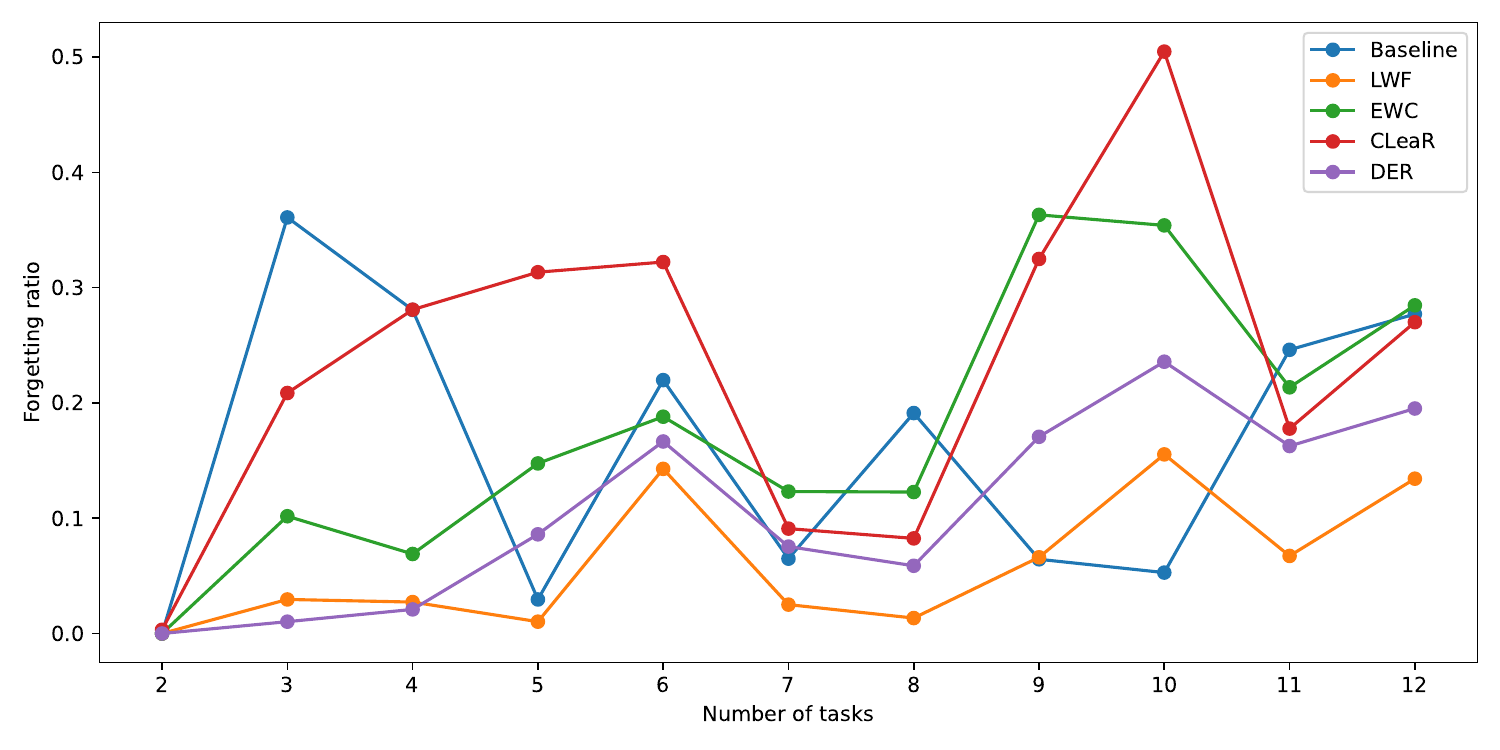}
    \caption{Changes of forgetting ratio ($F_{r}$) as the number of sequential tasks varies}
    \label{fig:fr}
\end{figure}

Fig.~\ref{fig:fr} illustrates how $F_{r}$ changes as each estimator is incrementally trained on a series of tasks. For example, when $K=8$, each estimator has been trained on eight tasks (T1 to T8 in Fig.~\ref{fig:tasks}). Then, $F_{r}$ is computed by performing the eight tasks with test datasets. As depicted in Fig.~\ref{fig:fr}, the baseline's $F_{r}$ is the highest among all estimators when $K=8$. This means that the baseline suffers from forgetting phenomena the most. LWF's $F_{r}$ is the lowest when $K=8$; it is least affected by forgetting phenomena. 

LWF and DER consistently keep their $F_{r}$ low as $K$ changes. When $K=2, 5, 6, 7, 8, 9, 11 \text{ and } 12$, LWF yields the lowest $F_{r}$ among all estimators. Its standard deviation of $F_{r}$ is 0.0559, and DER's is 0.0984. On the contrary, when $K=3, 4, 6, 8 \text{ and } 11$, the baseline yields the highest $F_{r}$ among all estimators. Its standard deviation of $F_{r}$ is 0.1489. 

In summary of the previous experimental results, LWF and DER stand out of all estimators and consistently outperform the baseline. LWF is preferable when estimation accuracy and $F_{r}$ are important factors to consider. However, it incurs higher computational cost than DER. DER is more suitable when required to balance the three factors (estimation accuracy, $F_{r}$ and computational cost). 

This paper also evaluates the ability of each estimator to perform {\it indirect} learning. For this evaluation, an estimator is tested on an unexperienced task that is similar to the previously learned tasks, even though it has not been trained on the new task. For example, after being trained on T1, T2 and T3 in Fig.~\ref{fig:tasks}, an estimator is tested for T4, which is unexperienced but similar to T1 to T3.  Table \ref{tab:indirect-learning} shows how much the test error decreases in an unexperienced task after being training on prior three tasks:  $\mathcal{L_\text{mse}}_{3}^{k} - \mathcal{L_\text{mse}}_{4}^{k}$. The higher this MSE difference is, the higher improvement of estimation accuracy an estimator produces. 

In Table \ref{tab:indirect-learning}, ``training sequence'' means a sequence of the tasks that an estimator has been trained on. ``Target task'' means an unexperienced task that the estimator is being tested for. After the training on T1, T2 and T3, the baseline decreases the test error on T4 by 0.079: $\mathcal{L_\text{mse}}_{3}^{k} - \mathcal{L_\text{mse}}_{4}^{k} = 0.079.$ DER produces a higher improvement: 0.14. DER and EWC consistently outperforms the baseline in all four training sequences. LWF and CLeaR outperforms the baseline in two sequences. Table~\ref{tab:indirect-learning} shows that CL-based estimators can approach unexperienced tasks well by leveraging the knowledge on the previously learned similar tasks.   

\begin{table}[htb]
    \centering
    \caption{MSE changes in indirect learning}
    \begin{tabular}{|c|c|| c | c | c | c | c|}
    \hline
    Training sequence & Target task & Baseline & LWF & EWC & CLeaR & DER \\
    \hline
    \hline
    $\text{T1} \rightarrow \text{T2} \rightarrow \text{T3} $ & \text{T4} & 0.079 & 0.045 & 0.087 & 0.200 & 0.140 \\
    $\text{T1} \rightarrow \text{T4} \rightarrow \text{T3} $ & \text{T2} & 0.020 & 0.075 & 0.110 & 0.017 & 0.080 \\
    $\text{T1} \rightarrow \text{T2} \rightarrow \text{T4} $ & \text{T3} & 0.050 & 0.023 & 0.137 & 0.195 & 0.229 \\
    $\text{T1} \rightarrow \text{T4} \rightarrow \text{T2} $ & \text{T3} & 0.065 & 0.118 & 0.163 & 0.035 & 0.327 \\
    \hline
    \end{tabular}
    \label{tab:indirect-learning}
\end{table}

\section{Conclusion}

This paper demonstrates the impacts of continual learning (CL) on the incremental performance estimation tasks for a feedback-based molecular communication protocol. CL allows performance estimators to continuously learn a series of new estimation tasks under unexperienced settings without compromising those that have been learned in the past. The proposed CL-based estimators can enhance a traditional neural network by improving estimation accuracy at a variety of computational costs. Findings of this paper can be applied to many other incremental estimation/prediction tasks in the field of molecular communication. 

\bibliographystyle{splncs04}
\bibliography{bict25}

@article{casaleiro2024sync,
  author={Casaleiro, Duarte and Souto, Nuno M. B. and Silva, João C.},
  journal={IEEE Access}, 
  title={Synchronization and Detection in Molecular Communication Using a Deep-Learning-Based Approach}, 
  year={2024},
  volume={12},
  number={},
  pages={192539-192553}
}

@article{cheng2025localtrack,
  author={Cheng, Zhen and Liu, Heng and Zheng, Jianlong and Gong, Weihua and Chi, Kaikai},
  journal={IEEE Sensors Journal}, 
  title={Localizing and Tracking the Transmitter Bionanosensor in Mobile Molecular Communication by Deep Learning}, 
  year={2025},
  volume={25},
  number={7},
  pages={10583-10593}
}

@article{kara2022idx,
    author = {Kara, Ozgur and Yaylali, Gokberk and Pusane, Ali and Tugcu, Tuna},
    title = {Molecular index modulation using convolutional neural networks},
    journal = {Nano Communication Networks},
    volume = {34},
    pages = {100420},
    year = {2022},
}

@inproceedings{yilmaz2017ml,
  author={Yilmaz, H. Birkan and Lee, Changmin and Cho, Yae Jee and Chae, Chan-Byoung},
  booktitle={Proc. IEEE Int'l Black Sea Conference on Communications and Networking}, 
  title={A machine learning approach to model the received signal in molecular communications}, 
  year={2017},
}

@inproceedings{lee2017mlchannel,
  author={Lee, Changmin and Yilmaz, H. Birkan and Chae, Chan-Byoung and Farsad, Nariman and Goldsmith, Andrea},
  booktitle={Proc. IEEE Int'l Workshop on Signal Processing Advances in Wireless Communications}, 
  title={Machine learning based channel modeling for molecular MIMO communications}, 
  year={2017},
}

@inproceedings{cheng2025bio,
  author={Cheng, Zhen and Liu, Heng and Chen, Miaodi and Zhang, Zhichao},
  title={Machine Learning-Based Detection Time Estimation for Molecular Communication},
  booktitle={Proc. Int'l Conf. on Bio-inspired Information and Communications Technologies},
  year={2024}
}

@article{cheng25dl,
  author={Cheng, Zhen and Liu, Heng and Xu, Ziyan and Li, Jiaxin and Chi, Kaikai},
  journal={IEEE Trans. Mol. Biol. Multi-Scale Commun.}, 
  title={Deep Learning-based Estimation of Emission Time and Arrival Time in Diffusive Multi-Receiver Molecular Communication}, 
    pages={early access},
  year={2025},
}

@article{vandeven22incremental,
	author={van de Ven, G.M. and Tuytelaars, T. and Tolias, A.S.},
	title={Three types of incremental learning},
	journal={Nat. Mach. Intell.},
	volume={4},
	pages={1185–1197},
	year={2022}
}

@article{wang24cl,
	author={Wang, Liyuan and Zhang, Xingxing and Su, Hang and Zhu, Jun},
	title={A Comprehensive Survey of Continual Learning: Theory, Method and Application}, 
	journal={IEEE Trans. Pattern Anal. Mach. Intell.}, 
	volume={46},
	number={8},
	pages={5362-5383},
	year={2024}
}

@article{mccloskey89catastrophic,
	author={McCloskey, M. and Cohen, N.J.},
	title={Catastrophic interference in connectionist networks: The sequential learning problem},
	journal={Psychology of Learning and Motivation},
	volume={24},
	pages={109-165},
	year={1989}
}

@inproceedings{hinton2015distillingknowledgeneuralnetwork,
      title={Distilling the Knowledge in a Neural Network},
      author={Geoffrey Hinton and Oriol Vinyals and Jeff Dean},
      booktitle={Proc. of NIPS 2014 Deep Learning Workshop},
      year={2024}
}

@inproceedings{buzzega20der,
	author = {Buzzega, Pietro and Boschini, Matteo and Porrello, Angelo and Abati, Davide and Calderara, Simone},
	title = {Dark experience for general continual learning: a strong, simple baseline},
	year = {2020},
	booktitle={Proc. Int'l Conference on Neural Information Processing Systems},
}

@article{li18lwf,
	title={Learning without Forgetting},
	author={Zhizhong Li and Derek Hoiem},
	journal={IEEE Trans. Pattern Anal. Mach.},
	volume={40},
	number={12},
	pages={2935-2947}, 
	year={2018}
}

@article{kirkpatrick17ewc,
	author={J. Kirkpatrick and R. Pascanu and N. Rabinowitz and et al.},
	title={Overcoming catastrophic forgetting in neural networks},
	journal={Proc. Natl. Acad. Sci. USA},
	volume={114},
	number={13},
	pages={3521-3526}, 
	year={2017}
}

@article{he21clear,
	author={He, Y. and Sick, B.},
	title={{CL}ea{R}: An adaptive continual learning framework for regression tasks},
	journal={AI Perspect},
	volume={3},
	number={2},
	year={2021}
}

@article{singh23ecc,
	title={{VLSI} implementation of error correction codes for molecular communication},
	author={Singh, S. P. and Rai, R. and Awasthi, S. and Singh, D. K. and Lakshmanan, M.},
	journal={Wirel. Personal. Commun.},
	volume={130},
	pages={2697-2713},
	year={2023}
}

@inproceedings{mitzman15swarq,
	author={J. S. Mitzman and B. Morgan and T. M. Soro and J. Suzuki and T. Nakano},
	title={A Feedback-based Molecular Communication Protocol for Noisy Intrabody Environments}, 
	booktitle={17th IEEE Int'l Conference on E-health Networking Applications and Services},
	year={2015},
}

@article{vale96nature, 
	author={R. D. Vale and T. Funatsu and D. W. Pierce and L. Romberg and Y. Harada and T. Yanagida},
	title={Direct Observation of Single Kinesin Molecules Moving along Microtubules},
	journal={Nature},
	volume={380},
	page={451-453},
	year={1996},
}

@article{burton72error,
	author={H. O. Burton and D. D. Sullivan},
	title={Errors and Error Control},
	journal={Proc. of the IEE},
	volume={60},
	number={11},
	year={1972},
}

@article{bai15arq-bacteria, 
	author={C. Bai and M. S. {Leeson et al.}},
	title={Performance of {SW-ARQ} in Bacterial Quorum Communications},
	journal={Nano Commun. Netw.},
	volume={6(1)}, 
	year={2015},
}

@article{wang14arq, 
	author={X. Wang and M. D. Higgins and M. S. Leeson},
	title={Simulating the Performance of {SW-ARQ} Schemes within Molecular Communications},
	journal={Simulation Modelling Practice and Theory},
	volume={42}, 
	page={178-188},
	year={2014},
}

@article{felicetti14tcpmol,
	author={L. Felicetti and M. Femminella and G. Reali and T. Nakano and A. V. Vasilakos},
	title={{TCP}-like molecular communications},
	journal={IEEE J. Sel. Area Comm.}, 
	volume={32},
	number={12}, 
	pages={2354-2367},
	year={2014},
}

@article{nakano13jsac,
	author={T. Nakano and Y. Okaie and A. V. Vasilakos},
	title={Transmission Rate Control for Molecular Communication among Biological Nanomachines},
	journal={IEEE J. Sel. Area Comm.}, 
	volume={31}, 
	number={12},
	pages={835-846},  
	year={2013},
}
\end{document}